\documentclass[10pt,twocolumn,letterpaper]{article}

\usepackage[pagenumbers]{iccv} 

\usepackage{times}
\usepackage{epsfig}
\usepackage{amsmath}
\usepackage{amssymb}
\usepackage[linesnumbered,ruled,vlined]{algorithm2e}
\usepackage{multibib}
\usepackage{xcolor,colortbl}
\usepackage{amsthm}

\usepackage{minitoc}
\usepackage{tocloft}

\usepackage{color}
\usepackage{booktabs}   
\usepackage{multirow}   
\usepackage{array}      
\usepackage{paralist}   
\usepackage{caption}    
\usepackage{tabularx}  
\definecolor{turquoise}{cmyk}{0.65,0,0.1,0.3}
\definecolor{purple}{rgb}{0.65,0,0.65}
\definecolor{dark_green}{rgb}{0, 0.5, 0}
\definecolor{orange}{rgb}{0.8, 0.6, 0.2}
\definecolor{red}{rgb}{0.8, 0.2, 0.2}
\definecolor{darkred}{rgb}{0.6, 0.1, 0.05}
\definecolor{blueish}{rgb}{0.3, 0.3, .6}
\definecolor{light_gray}{rgb}{0.7, 0.7, .7}
\definecolor{pink}{rgb}{1, 0, 1}
\definecolor{greyblue}{rgb}{0.25, 0.25, 1}
\definecolor{awesome}{rgb}{1.0, 0.13, 0.32}

\definecolor{figred}{rgb}{0.9, 0.1, 0.1}
\definecolor{figgreen}{rgb}{0.1, 0.7, 0.1}
\definecolor{figblue}{rgb}{0.1, 0.1, 0.9}
\definecolor{figmagenta}{rgb}{0.8, 0.1, 0.8}

\usepackage{blindtext}

\renewcommand{\paragraph}[1]{\vspace{1em}\noindent\textbf{#1}}

\usepackage{gensymb}

\usepackage{graphicx}               

\usepackage{tabu}

\definecolor{iccvblue}{rgb}{0.21,0.49,0.74}
\usepackage[pagebackref,breaklinks,colorlinks,allcolors=iccvblue]{hyperref}

\def\confName{ICCV}
\def\confYear{2025}

\usepackage{enumitem}
\usepackage{algorithmic}
\usepackage{graphicx}
\usepackage{multirow}
\usepackage{booktabs}

\def\ie{\emph{i.e., }}
\def\eg{\emph{e.g., }}

\title{
Adams Bashforth Moulton Solver for Inversion and Editing in Rectified Flow
}

\author{%
  Yongjia Ma$^1$\qquad 
    Donglin Di$^1$ \qquad     
  Xuan Liu$^2$ \qquad  
  Xiaokai Chen$^2$ \\[1ex]
    Lei Fan$^3$\qquad
    Tonghua Su$^2$\qquad
    Yue Gao$^4$\\[1ex]
  \\
    $^1$Li Auto\quad
    $^2$Harbin Institute of Technology\quad
    $^3$University of New South Wales\quad
    $^4$Tsinghua University\quad
}

\begin{document}

\twocolumn[{%
\renewcommand\twocolumn[1][]{#1}%
\maketitle
}]

\abstract{

Rectified flow models have achieved remarkable performance in image and video generation tasks. However, existing numerical solvers face a trade-off between fast sampling and high-accuracy solutions, limiting their effectiveness in downstream applications such as reconstruction and editing. To address this challenge, we propose leveraging the Adams-Bashforth-Moulton (ABM) predictor-corrector method to enhance the accuracy of ODE solving in rectified flow models.
Specifically, we introduce ABM-Solver, which integrates a multi-step predictor–corrector approach to reduce local truncation errors and employs Adaptive Step Size Adjustment to improve sampling speed. 
Furthermore, to effectively preserve non-edited regions while facilitating semantic modifications, we introduce a Mask Guided Feature Injection module. We estimate self-similarity to generate a spatial mask that differentiates preserved regions from those available for editing.
Extensive experiments on multiple high-resolution image datasets validate that ABM-Solver significantly improves inversion precision and editing quality, outperforming existing solvers without requiring additional training or optimization. 
}


\section{Introduction}
\label{sec:intro}

Diffusion models have emerged as the dominant paradigm for high-fidelity image and video generation \cite{ho2020denoising,song2020denoising,esser2024scaling,ramesh2022hierarchicaltextconditionalimagegeneration}. These models learn to reverse the forward process of transforming data into noise, effectively capturing high-dimensional perceptual data structures. 
Early approaches like DDPM \cite{ho2020denoising} employ stochastic differential equations (SDEs) for iterative denoising, which is computationally expensive during generation.
To enhance efficiency and accuracy, recent advancements in flow matching \cite{liu2022flow,lipman2022flow}  have demonstrated superior generative performance with faster training and inference \cite{flux2023,esser2024scaling}. This shift is attributed to architectures like DiT \cite{peebles2023scalable}, gradually replacing the UNet backbone \cite{rombach2022high} and reformulating SDEs as ordinary differential equations (ODEs). By transitioning from stochastic to deterministic formulations, these models directly learn smooth velocity fields, enabling faster and more stable sampling \cite{kim2024simple,lee2025improving,yan2025perflow}
. 

Building on earlier advancements, Rectified Flow (ReFlow) models optimize velocity trajectories by leveraging numerical solvers, such as the Euler method or high-order Taylor expansions to enforce near-linear transformations \cite{wang2024taming,avrahami2024stable}. These methods reduce cumulative errors and improve inversion quality. However, increasing the solver's order typically comes at the expense of computational efficiency; alternative approaches like the Midpoint Method \cite{deng2024fireflow}, while potentially faster, can be overly sensitive to step-size selection and may suffer from stability issues. Consequently, the inversion process becomes progressively less accurate over extended integration horizons, which directly undermines the performance of downstream applications including reconstruction and editing.

To address discretization errors arising from fixed-step integration, we present the ABM-Solver, a linear multistep method that integrates a two-step predictor–corrector approach \cite{durran1991third}, with an adaptive step size adjustment mechanism to dynamically adjust step sizes driven by local truncation error estimates. During the first stage (\ie, explicit prediction), the Adams–Bashforth method estimates the next state from a weighted sum of previously computed gradients, providing fast yet potentially coarse approximations. Subsequently, This is refined by the second stage (\ie, implicit corrector) via the Adams–Moulton method, which incorporates newly evaluated gradients to reduce local errors and improve stability. 

In addition to enhancing inversion precision and reconstruction accuracy, we thoroughly analyze the spatial consistency of attention values between the inversion and sampling stages. Specifically, we introduce a Mask Guided Feature Injection module that effectively supports a wide range of editing tasks while preserving the original content’s semantics. To achieve this, we estimate self-similarity to generate a spatial mask that differentiates between regions requiring preservation and those amenable to modification. This mask-guided fusion of attention features helps retain critical structural details in non-edited areas while enabling flexible semantic alterations where needed. Extensive theoretical analysis and empirical evaluations on multiple high-resolution image datasets demonstrate that our approach effectively balances computational efficiency and numerical precision, outperforming existing methods without requiring additional training or optimization costs. Our contributions are summarized as follows: 

\begin{figure}[t]
    \centering
    \includegraphics[width=\linewidth]{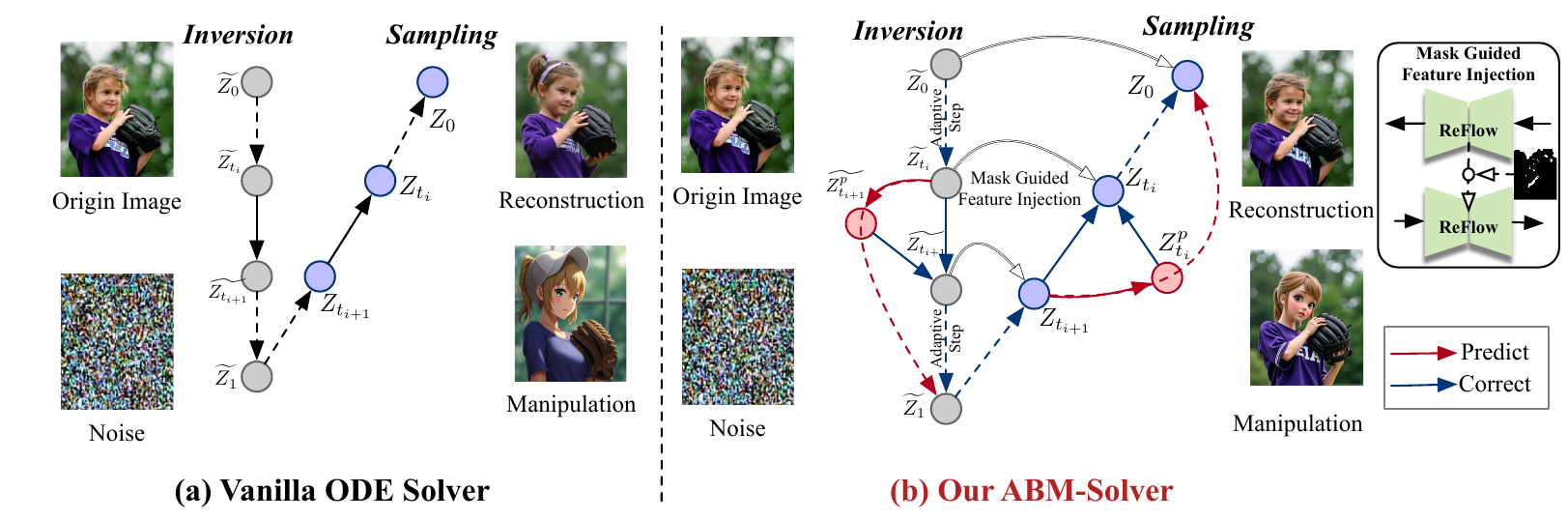}
    \caption{Comparison of Vanilla ODE Solver and Our ABM-Solver. (a)\textbf{ Vanilla ODE Solver}: Conventional numerical integration accumulates approximation errors, compromising inversion accuracy and editing fidelity. (b) \textbf{Our ABM-Solver}: By integrating a second-order ABM predictor–corrector framework, the Adaptive Step Size Adjustment, and Mask Guided Feature Injection, our method significantly reduces errors and achieves superior inversion and semantic editing performance.}
    \label{fig:enter-label}
\end{figure}

\begin{itemize}

\item We propose ABM-Solver, a training-free numerical solver that integrates the Adams–Bashforth–Moulton predictor–corrector method with Adaptive Step Size Adjustment, achieving second-order accuracy and reduced local truncation errors in Rectified Flow models without extra computational cost.
\item We introduce a Mask Guided Feature Injection module, which leverages spatial consistency of attention values to generate masks for selectively fusing features, thereby preserving structural details while enabling semantically consistent editing. 
\item Extensive theoretical analysis and experiments on high-resolution image datasets demonstrate that our approach significantly improves inversion precision and reconstruction accuracy, and supports diverse editing tasks, outperforming existing solvers without requiring extra training. 
\end{itemize}


\section{Related Work}

\subsection{Diffusion Inversion}
Diffusion inversion maps the real image back to the Gaussian noise to recover its corresponding initial noise \cite{song2020denoising, 10.5555/3737916.3739382, zhang2025easyinvfastbetterddim, delbracio2024inversiondirectiterationalternative}. This process is crucial for a wide range of tasks, including image reconstruction and editing \cite{yang2024tv3dgmasteringtextto3dcustomized,lin2024scheduleeditsimpleeffective,brack2024leditslimitlessimageediting,deutch2024turboedittextbasedimageediting,couairon2022diffeditdiffusionbasedsemanticimage, cho2024noisemapguidanceinversion}. 
However, the reconstruction fidelity of DDIM inversion is notably compromised when using a limited number of denoising steps, especially in the context of text-guided diffusion models \cite{hertz2022prompt,samuel2025lightningfastimageinversionediting}.
To achieve more accurate reconstruction, noise map guidanc \cite{cho2024noise} employs a gradient from the noise map to direct the reverse diffusion process, forcing its path to correspond with the forward diffusion trajectory.
Null-Text Inversion (NTI) \cite{mokady2023null} achieves high-fidelity reconstruction by fine-tuning the null-text embedding for each image, but this per-instance optimization strategy comes at a significant computational cost.
NPI \cite{miyake2023negative} trades reconstruction accuracy for computational efficiency by employing a static negative prompt embedding to bypass the optimization step.
ReNoise \cite{garibi2024renoiserealimageinversion} enhances reconstruction fidelity by employing the backward Euler (or implicit Euler) method as its numerical solver for the inversion process.
FreqInv \cite{bao2025freeinvfreelunchimproving} enhances reconstruction by performing inversion in the frequency domain, where it treats low-frequency structure and high-frequency texture components distinctly. However, this signal decomposition, while effective, renders the process computationally intensive.

\subsection{Reflow Inversion}
Rectified flow models aim to straighten the trajectory between noise and data, offering a more direct path for inversion \cite{liu2022flow,kim2024simple}. 
RF-Inversion \cite{rout2024semantic} frames the task as a dynamic optimal control problem, deriving a controller via a Linear Quadratic Regulator (LQR). It constructs a new vector field that optimally balances fidelity to the input image against consistency with the model's learned distribution, thereby improving inversion accuracy without requiring costly test-time optimization. 
RF-Solver \cite{wang2024taming} applies high-order Taylor expansions to reduce approximation errors in the ODEs of rectified flow. FireFlow \cite{deng2024fireflow} introduces a numerical solver that achieves second-order precision with the computational cost of a first-order solver. EE-Edit \cite{yan2025eeditrethinkingspatialtemporal} achieves significant inversion acceleration by exploiting spatial and temporal redundancy through feature caching and inversion step skipping, however, its core caching mechanism is complex and introduces additional computational and memory overhead.
However, these methods either require increased computational resources or struggle to balance efficiency and precision in inversion accuracy.

\subsection{Editing with Inversion Methods} 
 Current editing methods \cite{duan2023tuningfreeinversionenhancedcontrolconsistent,chen2024controlavideocontrollabletexttovideodiffusion,kara2023raverandomizednoiseshuffling,brooks2023instructpix2pixlearningfollowimage} are fundamentally built upon accurate inversion, which transforms a real image into a latent representation within the model's generative space, making it amenable to precise manipulation.
Prompt-to-Prompt (P2P) editing \cite{hertz2022prompt}, directly controls image synthesis by strategically altering the cross-attention maps; specifically, it replaces or re-weights the attention scores tied to words in the source prompt with those from the target prompt. 
 To better preserve spatial layout during complex transformations, MasaCtrl \cite{cao2023masactrl} introduces a mutual self-attention mechanism, where the generation process for the edited image queries the self-attention layers from the source inversion to maintain consistent structure and pose in non-edited regions.  Plug-and-Play (PnP) methods \cite{tumanyan2023plug} fuse the spatial layout from the source image into the edited output by injecting the source's self-attention features directly into the denoising process of the target image.
Similarly, Pix2Pix-Zero \cite{parmar2023zero} employs cross-attention guidance to maintain structural consistency, while uniquely eliminating the need for per-image prompting by using pre-computed editing directions.
Despite these advancements, existing methods \cite{lu2022dpmsolverfastodesolver,zhang2023addingconditionalcontroltexttoimage,song2020denoising} either require many inversion steps, leading to increased computational time or struggle to balance editing flexibility with preservation of the source image's content. 
By addressing these limitations, our proposed ABM-Solver provides a more accurate inversion with fewer steps due to its higher-order accuracy and theoretical convergence properties, enhancing the effectiveness of editing methods with minimal computational costs.


\vspace{-6pt}

\section{Preliminaries}
\vspace{-6pt}

In this section, we provide the necessary background on Rectified Flow models and their mathematical formulations, which form the foundation of our proposed method. Flow-based generative models learn to map a simple prior distribution \(\pi_1\) (typically Gaussian noise) to a complex data distribution \(\pi_0\) (\eg, images). 

Given samples \(\mathbf{Z}_0 \sim \pi_0\) and \(\mathbf{Z}_1 \sim \pi_1\), the goal is to learn a transformation that maps \(\mathbf{Z}_0\) to \(\mathbf{Z}_1\) through a continuous flow over time \(t \in [0, 1]\). The transformation is represented by an ordinary differential equation (ODE):
\begin{equation}
\frac{d\mathbf{Z}(t)}{dt} = \mathbf{v}_{\theta}(\mathbf{Z}(t), t),
\label{eq:ode}
\end{equation}
where\(\mathbf{v}_{\theta}: \mathbb{R}^d \times [0, 1] \rightarrow \mathbb{R}^d\) represents a time-dependent vector field, learned by a neural network with parameters $\theta$. The velocity field \(\mathbf{v}_{\theta}\) governs the flow of the latent variables over time, guiding the transformation from the source distribution (at $t=0$) to the target (at $t=1$). \(\mathbf{Z}(t) \in \mathbb{R}^d\) is the state at time \(t\).

\subsection{Forward Process}
Rectified Flow (ReFlow) \cite{liu2022flow} offers an elegant solution by postulating that the most direct path between a noise sample \(\mathbf{Z}_1 \sim \pi_1\) and a data sample \(\mathbf{Z}_0 \sim \pi_0\) is a straight line. This "rectified" trajectory is simply a linear interpolation:
\begin{equation}
\mathbf{Z}_t = (1 - t) \mathbf{Z}_0 + t \mathbf{Z}_1, \quad t \in [0, 1],
\label{eq:linear_interpolation}
\end{equation}


\noindent The time derivative of this linear trajectory defines a non-causal and simple ordinary differential equation (ODE):
\begin{equation}
\frac{d\mathbf{Z}_t}{dt} = \mathbf{Z}_1 - \mathbf{Z}_0.
\end{equation}
However, this formulation requires knowledge of \(\mathbf{Z}_1\) at all intermediate time steps, which is impractical for simulation. To overcome this, a velocity field \(\mathbf{v}_{\theta}(\mathbf{Z}(t), t)\) is learned to approximate the linear interpolation direction without requiring \(\mathbf{Z}_1\) during the forward process. 
The causalized forward ODE is given by Equation~\eqref{eq:ode}.

To create a practical, causal model, ReFlow trains a neural network \(\mathbf{v}_{\theta}(\mathbf{Z}_t, t)\) to approximate this ideal velocity. The network is optimized to predict the direction \(\mathbf{Z}_1 - \mathbf{Z}_0\) given only an intermediate point \(\mathbf{Z}_t\) on the straight path. This is achieved by minimizing the L2 distance between the predicted and ideal velocities, averaged over all possible paths and times:
\begin{equation}
\min_{\theta} \mathbb{E}_{(\mathbf{Z}_0, \mathbf{Z}_1), t} \left[ \left\| \mathbf{v}_{\theta}(\mathbf{Z}_t,t) - (\mathbf{Z}_1 - \mathbf{Z}_0) \right\|_2^2 \right].
\label{eq:reflow_objective}
\end{equation}
This training objective forces the learned dynamics to adhere closely to straight-line trajectories, simplifying the flow and improving model stability.




\subsection{Reverse Process}

Once the forward process has been trained to map samples from the source distribution \(\pi_0\) to the target distribution \(\pi_1\), generating new samples from the target distribution requires the reverse process. The goal of the reverse process is to transform samples from the target distribution \(\pi_1\) back to the source distribution \(\pi_0\), essentially "reversing" the trajectory of the forward process.
Starting with a sample \(\mathbf{Z}(1) \sim \pi_1\) from the target distribution at time \(t=1\), the reverse process is governed by the following ordinary differential equation (ODE):
\begin{equation}
\frac{d\mathbf{Z}(t)}{dt} = -\mathbf{v}_{\theta}(\mathbf{Z}(t), t), \quad t \in [1, 0],
\label{eq:ode_backward}
\end{equation}
where \(\mathbf{v}_{\theta}(\mathbf{Z}(t), t)\) represents the velocity field learned during the forward process, and the negative sign indicates that the dynamics are reversed. The reverse ODE effectively undoes the transformation that was applied in the forward process by gradually moving backwards through the learned flow.

\vspace{-6pt}

\section{Methods}
\vspace{-6pt}

\subsection{Adams-Bashforth-Moulton Sovler}
\vspace{-6pt}
RF-based models are capable of generating high-quality images and videos. However, when applied to reconstruction and editing tasks, recent studies have found that approximations in solving the corrected flow ODE lead to significant error accumulation at each time step \cite{wang2024taming,deng2024fireflow}. Therefore, by improving the numerical integration process to obtain a more accurate solution to the ODE, we aim to reduce error accumulation, thereby enhancing the performance of corrected flow models in inversion and editing tasks.

Based on this analysis, we first carefully examine the differential form of the corrected flow: \( d\mathbf{Z}(t) = -\mathbf{v}_{\theta}(\mathbf{Z}(t), t) \, dt \). During the sampling process, this ordinary differential equation (ODE) is discretized. Given the initial value \(\mathbf{Z}_{t_i}\), the ODE can be accurately represented using the variant of constant method:
\begin{equation}
    \label{eq:integral_zti_ode}
    \mathbf{Z}_{t_{i-1}} 
    \;=\;
    \mathbf{Z}_{t_{i}} 
    \;+\;
    \int_{t_i}^{t_{i-1}}
    \mathbf{v}_{\theta}\bigl(\mathbf{Z}(\tau), \tau\bigr)\, d\tau,
\end{equation}
where $\mathbf{Z}(\tau)$ denotes the evolving latent state (\eg, an image tensor) and $\mathbf{v}_{\theta}$ is a neural-network-based velocity field designed for mapping between noise and real data distributions via a straight-line flow. A multi-step method approximates the integral by interpolating or extrapolating the function $\mathbf{v}_{\theta}(\mathbf{Z}(t_j), t_j)$ at several previous time points $t_j$.

\subsubsection{Predictor Step: Adams-Bashforth Method}

In the Adams-Bashforth method, the integral in Equation (\ref{eq:integral_zti_ode}) is approximated by an explicit polynomial extrapolation based on previously computed evaluations of the velocity field at earlier time steps. The main idea behind this method is to use the velocity values at several previous time points to predict the future state of the system.

Specifically, for a given time step \(t_i\) and step size \(h = t_{i-1} - t_i\), the two-step Adams-Bashforth method approximates the integral as:
\begin{equation}
\int_{t_i}^{t_{i-1}} \mathbf{v}_{\theta}(\mathbf{Z}(t), t) \, dt \approx h_i \sum_{j=0}^{p-1} \beta_j \mathbf{v}_\theta(\mathbf{Z}_{t_{i+j}}, t_{i+j}),
\label{eq:ab_approx}
\end{equation}
where \(\beta_j\) are the coefficients for the extrapolation, and the sum involves velocity field values at the previous time steps \(t_{i+j}\). The method relies on an explicit prediction from the previous steps, where the velocity field values at these steps guide the estimation of the integral over the current interval.

Once the integral approximation is computed, the predictor step computes an initial estimate for the system state at the next time step \(t_{i-1}\). This is achieved by adding the computed approximation to the current state \(\mathbf{Z}_{t_i}\):
\begin{equation}
\mathbf{Z}_{t_{i-1}}^{(p)} = \mathbf{Z}_{t_i} + \frac{h_i}{2} \left[ 3\,\mathbf{v}_{\theta}(\mathbf{Z}_{t_i}, t_i) - \mathbf{v}_{\theta}(\mathbf{Z}_{t_{i+1}}, t_{i+1}) \right].
\label{eq:ab2_predictor_en}
\end{equation}
This equation provides the prediction \(\mathbf{Z}_{t_{i-1}}^{(p)}\) of the system state at the next time step, based on the previous evaluations of the velocity field. Since the Adams-Bashforth method is explicit, it is straightforward to compute and can be parallelized, making it computationally efficient.

\subsubsection{Corrector Step: Adams-Moulton Method}

Once the predictor step provides an estimate \(\mathbf{Z}_{t_{i-1}}^{(p)}\) of the system state at \(t_{i-1}\), the Adams-Moulton method is applied to refine this estimate through implicit integration. The corrector step improves the prediction by incorporating additional information, particularly the velocity field at \(t_{i-1}\), which was not included in the explicit prediction.

The Adams-Moulton method employs the following integral approximation, which considers the velocity field at the predicted state \(t_{i-1}\) along with previous states:
\begin{equation}
\small
\begin{aligned}
\int_{t_i}^{t_{i-1}} \mathbf{v}_{\theta}(\mathbf{Z}(t), t) \, dt \approx
 h_i [ \alpha_0 \mathbf{v}_\theta(\mathbf{Z}_{t_{i-1}}^{(p)}, t_{i-1}) + 
 \sum_{j=0}^{p-1} \alpha_{j+1} \mathbf{v}_\theta(\mathbf{Z}_{t_{i+j}}, t_{i+j}) ],
\end{aligned}
\label{eq:am_approx}
\end{equation}
where \(\alpha_j\) are the coefficients for the extrapolation. This equation refines the integral estimate by including the velocity field evaluation at \(t_{i-1}\), which was initially ignored in the Adams-Bashforth step.

The corrected estimate of \(\mathbf{Z}_{t_{i-1}}\) is then computed by updating the state based on the refined approximation:
\begin{equation}
\small
\begin{aligned}
\mathbf{Z}_{t_{i-1}} 
&= \mathbf{Z}_{t_i} + \frac{h_i}{2} [ \,\mathbf{v}_{\theta}(\mathbf{Z}_{t_{i-1}}^{(p)}, t_{i-1}) + \,\mathbf{v}_{\theta}(\mathbf{Z}_{t_i}, t_i) ].
\end{aligned}
\label{eq:am2_corrector_en}
\end{equation}
This correction step improves the prediction obtained from the Adams-Bashforth method, leading to a more accurate estimate of the system's state at the next time step. The key advantage of the Adams-Moulton method is that it considers the velocity field at the predicted state, which helps improve the stability and accuracy of the solution.

To balance computational efficiency and accuracy, we employ the second-order Adams-Bashforth-Moulton (ABM) method. This method combines the explicit Adams-Bashforth method with the implicit Adams-Moulton method to provide an efficient and accurate numerical integration. The procedure consists of two steps: first, an explicit prediction step using the second-order Adams-Bashforth method, followed by an implicit correction step using the second-order Adams-Moulton method.

\subsubsection{Second-Order Runge-Kutta Initialization}
The ABM method requires starting values from previous time steps. To obtain accurate initial values, we use the second-order Runge-Kutta (RK2) method for the first step:
\begin{equation}
\begin{aligned}
    \mathbf{Z}_{t_{N-1}} = \mathbf{Z}_{t_N} + \frac{h}{2} ( \mathbf{v}_\theta(\mathbf{Z}_{t_N}, t_N) + \mathbf{v}_\theta(\mathbf{Z}_{t_N} + h \mathbf{v}_\theta(\mathbf{Z}_{t_N}, t_N), t_N + h) ).
\end{aligned}
\label{qe:rk2}
\end{equation}



\begin{algorithm}[t]
\caption{ABM-Solver for Rectified Flow ODE (Second-Order)}
\label{alg:abm_solver_second_order}
\begin{algorithmic}
\STATE Input: origin latent $\mathbf{Z_{}}$, time steps $\{t_N, t_{N-1}, \dots, t_0\}$

\STATE Initialize: \\

$\mathbf{k}_1 = \mathbf{v}_{\theta}(\mathbf{Z}_{t_N}, t_N)$ \\
$\mathbf{k}_2 = \mathbf{v}_{\theta}\left( \mathbf{Z}_{t_N} + h \mathbf{k}_1, t_N + h \right)$ \\
$\mathbf{Z}_{t_{N-1}} = \mathbf{Z}_{t_N} + \frac{h}{2} ( \mathbf{k}_1 + \mathbf{k}_2 )$

 \FOR{each timestep $t_i$ from $t_N$ to $t_0$}
    \STATE Predict: $\rightarrow$ Equation (\ref{eq:ab2_predictor_en})
    
    \STATE Correct: $\rightarrow$ performing Equation (\ref{eq:am2_corrector_en})
    
\ENDFOR
\STATE Output $\mathbf{Z}_{t_N}$
\end{algorithmic}
\end{algorithm}








\noindent The complete algorithm for the ABM-Solver is presented in Algorithm~\ref{alg:abm_solver_second_order}. This method significantly improves precision in reconstruction and editing tasks, particularly when using fewer timesteps.

Furthermore, consider an ODE \(\frac{d\mathbf{Z}}{dt} = v_\theta(\mathbf{Z}, t)\) with solution \(\mathbf{Z}(t)\). Let \(\Delta t\) be the step size, and assume \(v_\theta\) is sufficiently smooth in both \(\mathbf{Z}\) and \(t\). For a single step from \(t\) to \(t + \Delta t\), the two-step Adams-Bashforth-Moulton solver has a local truncation error of \(\mathcal{O}(\Delta t^3)\). More precisely, if \(\mathbf{Z}(t)\) is assumed exact at previous steps, then the difference between the true solution \(\mathbf{Z}(t+\Delta t)\) and the ABM numerical update \(\mathbf{Z}_{t+\Delta t}\) is on the order of \(\Delta t^3\). Consequently, over the entire interval \([0, T]\), the global error typically accumulates to \(\mathcal{O}(\Delta t^2)\), improving upon the \(\mathcal{O}(\Delta t)\) global error of a first-order (Euler) method:
\begin{equation}
    \begin{cases}
        \underbrace{\bigl\lVert \mathbf{Z}_{t+\Delta t} \;-\; \mathbf{Z}_{t+\Delta t}^{(\text{ABM})}\bigr\rVert}_{\text{local error}} 
\;\;\le\;\; C\,(\Delta t)^3,\\
\underbrace{\bigl\lVert \mathbf{Z}_t \;-\; \mathbf{Z}_{t}^{(\text{ABM})}\bigr\rVert}_{\text{global error}} 
\;\;\le\;\; C'\,(\Delta t)^2,
    \end{cases}
    \label{eq:error}
\end{equation}
\noindent analogous to the analysis in FireFlow~\cite{deng2024fireflow}, these error bounds hinge on \(v_\theta\) being Lipschitz-continuous in \(\mathbf{Z}\) and sufficiently smooth in \(t\). Under these conditions, the predictor step (Adams-Bashforth) and corrector step (Adams-Moulton) both incur only \(\mathcal{O}(\Delta t^3)\) local defects, leading to the overall \(\mathcal{O}(\Delta t^2)\) global accuracy. 

\vspace{-6pt}

\subsection{Adaptive Step Size Adjustment}
\label{subsec:adaptive}

To balance numerical accuracy and computational efficiency, we develop an Adaptive Step Size Adjustment strategy for the ABM solver. The key idea is to dynamically adjust the integration step size based on local truncation error estimates, allowing larger steps in regions where the solution varies smoothly and smaller steps near high-curvature regions.

Let $h_i = t_{i-1} - t_i$ denote the current step size, and define the local error estimate $\epsilon_i$ as the discrepancy between the predictor and corrector steps:
\begin{equation}
E_{i-1} = \left\| \mathbf{Z}_{t_{i-1}}^{(p)} - \mathbf{Z}_{t_{i-1}} \right\|_2,
\label{eq:error_estimate}
\end{equation}
where $\mathbf{Z}_{t_{i-1}}^{(p)}$ and $\mathbf{Z}_{t_{i-1}}$ are the predicted and corrected states from Equation.~(\ref{eq:ab2_predictor_en}) and (\ref{eq:am2_corrector_en}), respectively. The step size adaptation follows the error-controlled update rule:
\begin{equation}
h_{i-1}
\;=\;
h_i
\;\times\;
\left(\frac{\varepsilon}{E_{i-1}}\right)^{\!\!\frac{1}{p+1}},
\label{eq:step_adapt}
\end{equation}
where $\varepsilon$ is the error tolerance, and $p=2$ represents the method's order. This adaptation mechanism automatically reduces step sizes when local errors exceed the tolerance $\varepsilon$ while expanding steps in smooth regions where errors remain below the threshold. The default setting of $\varepsilon$ is 0.1.

It maintains numerical stability by preventing error accumulation through local error monitoring and improves computational efficiency. This dynamic balance makes the method particularly suitable for inversion and editing tasks where both precision and efficiency are critical.


\subsection{Mask Guided Feature Injection}

In the context of diffusion model-based image and video editing tasks, achieving a balance between semantic editing and preserving the structural integrity of the source image is of paramount importance. Notably, in the DiT model, the Attention Values ($V$) play a significant role in influencing the editing outcomes. To this end, by meticulously analyzing the spatial consistencies between the inversion and sampling processes with respect to the attention values, the regions amenable to editing can be more precisely identified, thereby striking an optimal balance between editability and fidelity to the input image.

\begin{figure}[t]
    \centering
    \includegraphics[width=\linewidth]{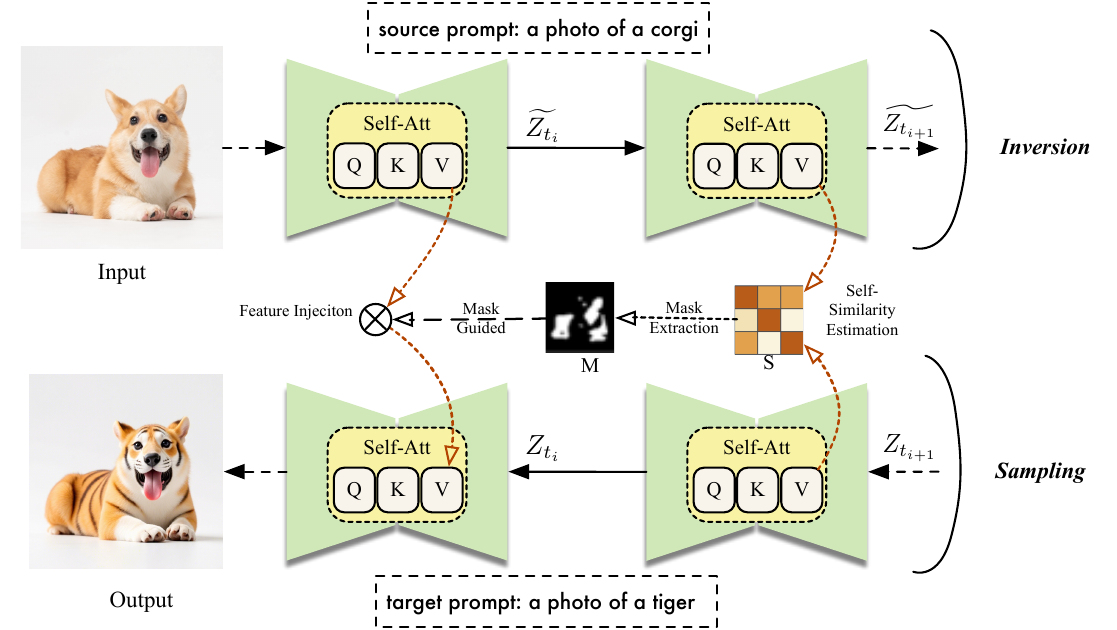}
    \caption{Overview of our Mask Guided Feature Injection module.
    Given the attention values from the inversion process $\widetilde{V}_{t_i}$ and the sampling process $V_{t_i}$, we first compute a pixel-wise cosine similarity map, which is then thresholded to generate a binary mask $\mathbf{M}$. 
    This mask identifies regions of high structural consistency. 
    It then guides the fusion of attention values for the subsequent timestep $t_{i-1}$, injecting features from the inversion process $\widetilde{V}_{t_{i-1}}$ into the preserved regions. 
    This ensures that the structural integrity of the source is maintained while allowing for targeted semantic edits.}
    \label{fig:Mask Guided Feature Injectionl}
    \vspace{-2pt}
\end{figure}

To reconcile the preservation of structural information with semantic editing, we conduct an in-depth analysis of the spatial consistencies between the attention values in the inversion and sampling processes. For aligned timesteps $t_i$, we define:
\vspace{-4pt}
\begin{equation}
\widetilde{V}_{t_i} = \mathcal{F}(\widetilde{Z}_{t_i}, t_i), \quad
V_{t_i} = \mathcal{F}(Z_{t_i}, t_i),
\end{equation}
where $\mathcal{F}$ represents the function mapping the latent variables and timestep to the corresponding attention values.


\noindent \textbf{Self Similarity Mask Estimation.} To quantitatively measure the similarity between the Attention Values in the inversion and sampling processes, we compute the pixel-wise cosine similarity $\mathbf{S}_{t_i}$. Cosine similarity is a commonly employed metric for gauging the directional alignment between two vectors. In our context, calculating the cosine similarity between $\widetilde{V}_{t_i}$ and $V_{t_i}$ determines the degree of similarity in the attention values at the same timestep across different processes. The computation is given by:
\begin{equation}
\mathbf{S}_{t_i} = \frac{\widetilde{V}_{t_i} \cdot V_{t_i}}{\|\widetilde{V}_{t_i}\| \|V_{t_i}\|}.
\end{equation}



\noindent We generate a spatial mask  $\mathbf{M}$ using a threshold $\tau$:
\begin{equation}
\mathbf{M} = 
\begin{cases}
1 & \mathbf{S}_{t_i}  \geq \tau \\
0 & \text{otherwise}
.
\end{cases}
\end{equation}
If the cosine similarity $\mathbf{S}$ at a pixel location meets or exceeds the threshold $\tau$, the corresponding value in the spatial mask $\mathbf{M}$ is set to $1$, signifying that the attention values in that region exhibit strong consistency between the inversion and sampling processes and should be retained. Conversely, if $\mathbf{S}$ is below $\tau$, the corresponding value in $\mathbf{M}$ is set to $0$, indicating that the region may be subject to editing. The default threshold $\tau$ is set to 0.2.

\noindent \textbf{Mask Guided Feature Fusion.} Once the spatial mask $\mathbf{M}$  is obtained, it is used to merge the attention values, ensuring the preservation of crucial features. Specifically, we adopt a mask-directed blending strategy to combine $\widetilde{V}_{t_{i-1}}$ and $V_{t_{i-1}}$, formulated as:
\begin{equation}
V_{t_{i-1}} = \mathbf{M} \odot\widetilde{V}_{t_{i-1}} + (1 - \mathbf{M}) \odot V_{t_{i-1}},
\end{equation}
where $\odot$ stands for element-wise multiplication. 

This strategy leverages the broader contextual estimation provided by the mask from the prior time step to mitigate localization errors in attention fusion. 
Consequently, during the editing procedure, we can successfully maintain the structural integrity of the source image while also fulfilling the requirements of semantic editing. Through this meticulous manipulation and examination of the Attention Values in both inversion and sampling phases, we gain enhanced control over the model's editing conduct. This allows us to preserve the original image structure and achieve high-caliber semantic editing, striking an excellent balance between editability and fidelity.

\section{Expriment}
\subsection{Setup}
\textbf{Baselines} This section compares Our ABM-solver with DM inversion-based editing methods such as Prompt-to-Prompt \cite{hertz2022prompt} , MasaCtrl \cite{cao2023masactrl}, Pix2Pixzero \cite{parmar2023zero}, Plug-and-Play \cite{tumanyan2023plug}, and DirectInversion \cite{elarabawy2022direct}. We also consider the recent RF inversion methods, such as RF-Inversion \cite{rout2024semantic} , RF-Solver \cite{wang2024taming} and FireFlow \cite{deng2024fireflow}.

\textbf{Metric} To comprehensively assess different methods, we conduct evaluations from three key perspectives: generation quality, preservation quality, and text-guided quality. Firstly, to evaluate the preservation quality of non-edited regions, a series of metrics come into play. These include the Learned Perceptual Image Patch Similarity (LPIPS) \cite{zhang2018unreasonable}, which measures certain perceptual similarities; the Structural Similarity Index Measure (SSIM) \cite{wang2004image} that focuses on structural likeness; the Peak Signal-to-Noise Ratio (PSNR) for quantifying signal and noise relationships; and also the structural distance \cite{heusel2017gans}. Lastly, in order to figure out the similarity between the generated image and the guiding text, a CLIP model \cite{radford2021learning} is utilized.

\subsection{Image Editing}
\vspace{-3pt}


We evaluate the ABM-Solver's image editing performance through tasks such as subject replacement, geometry changes, object addition, and global modifications. For object substitution and insertion, it is essential to preserve the background, while global edits like style transfer require maintaining the overall structure and composition of the original image. ABM-Solver meets these objectives, delivering high-quality results and preserving the integrity of the original content.

\begin{table*}[t]
\centering
\footnotesize
\caption{
Comparison of our ABM-Solver approach with other image editing methods on the PIE-Bench~\cite{ju2024pnp}. Metrics include background preservation (measured by PSNR and SSIM), structural consistency (measured by distance), and CLIP similarity for both the whole and edited images. Additionally, the number of steps and function evaluations (NFE) are provided to reflect the efficiency of each method.
}

\resizebox{\linewidth}{!}{

\begin{tabular}{l|c|c|c|c|c|c|c|c}
\toprule
\multirow{2}{*}{\textbf{Method}} & \multirow{2}{*}{\textbf{Model}} & \textbf{Structure} & \multicolumn{2}{c|}{\textbf{Background Preservation}} & \multicolumn{2}{c|}{\textbf{CLIP Similarity}$\uparrow$} & \multirow{2}{*}{\textbf{Steps}} & \multirow{2}{*}{\textbf{NFE}$\downarrow$} \\
\cmidrule(lr){3-4} \cmidrule(lr){4-5} \cmidrule(lr){6-7}
 & & \textbf{Distance}$\downarrow$ & \textbf{PSNR}$\uparrow$ & \textbf{SSIM}$\uparrow$ & \textbf{Whole} & \textbf{Edited} & & \\
\midrule
Prompt2Prompt \cite{hertz2022prompt} & Diffusion & 0.0694 & 17.87 & 0.7114 & 25.01 & 22.44 & 50 & 100 \\
Pix2Pix-Zero \cite{parmar2023zero} & Diffusion & 0.0617 & 20.44 & 0.7467 & 22.80 & 20.54 & 50 & 100\\
MasaCtrl \cite{cao2023masactrl} & Diffusion & 0.0284 & 22.17 & 0.7967 & 23.96 & 21.16 & 50& 100 \\ 
PnP \cite{ju2024pnp} & Diffusion & 0.0282 & 22.28 & 0.7905 & 25.41 & 22.55 & 50& 100\\
PnP-Inv. \cite{elarabawy2022direct} & Diffusion & 0.0243 & 22.46 & 0.7968 & 25.41 & 22.62 & 50 & 100 \\
\midrule
RF-Inversion \cite{rout2024semantic} & ReFlow & 0.0406 & 20.82 & 0.7192 & 25.20 & 22.11 & 28 & 56 \\
RF-Solver \cite{wang2024taming} & ReFlow & 0.0311 & 22.90 & 0.8190 & {26.00} & {22.88} & 15 & 60 \\
FireFlow \cite{deng2024fireflow} & ReFlow & 0.0283 & {23.28} & {0.8282} & 25.98 & {22.94} & 15 & 32 \\


\textit{Ours (ABM-Solver)} & ReFlow & \textbf{0.0207} & \textbf{24.60} & \textbf{0.8305} & \textbf{27.31} & \textbf{22.97} & {15} & {40-60} \\

\bottomrule
\end{tabular}
}
\label{tab:editing}
\vspace{-4pt}
\end{table*}

\begin{figure*}[t]
    \vspace{-12pt}

    \centering
\includegraphics[width=\linewidth]{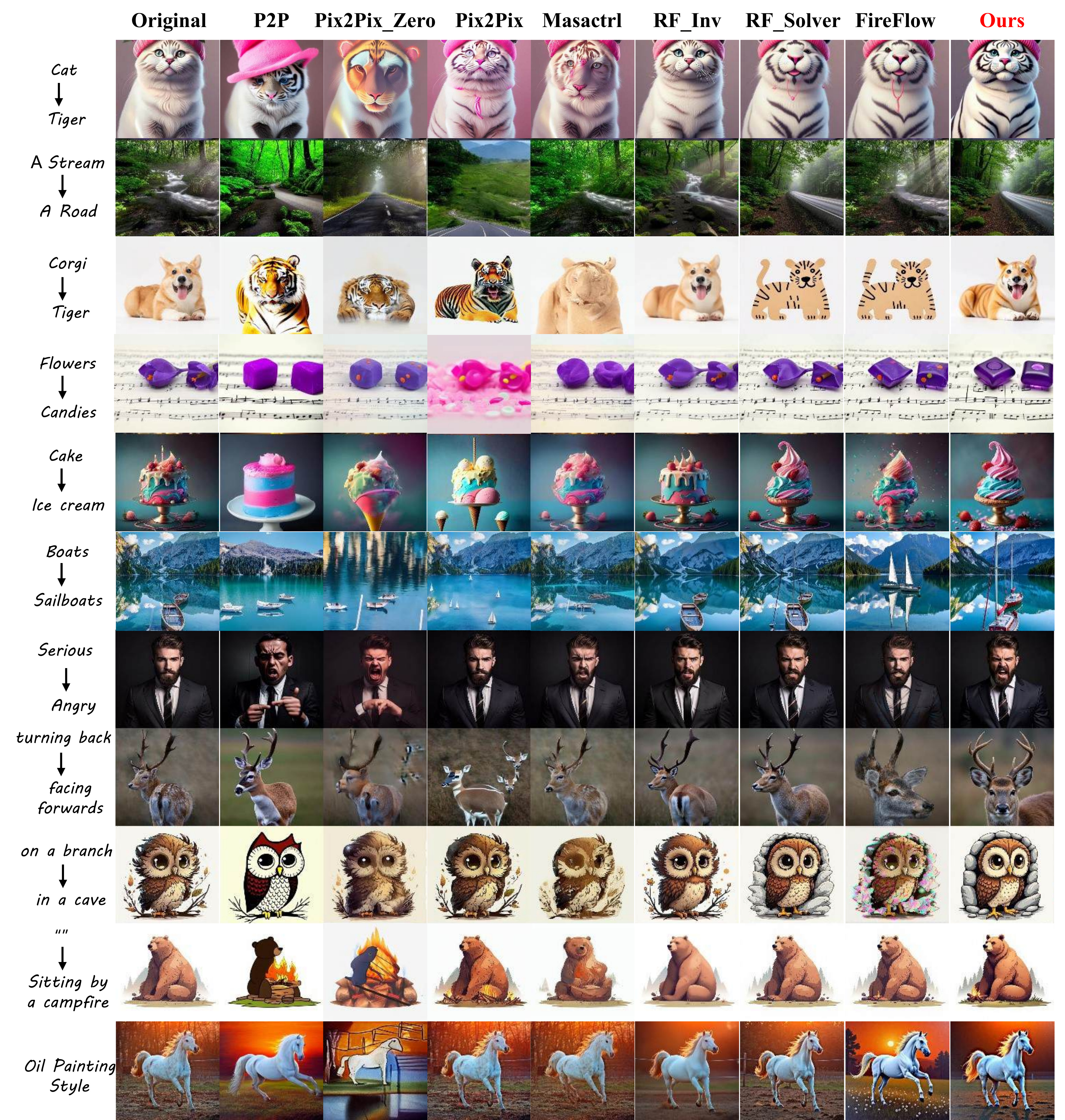}
    \caption{Qualitative comparison of image editing results across different methods. Our ABM-Solver demonstrates superior performance, delivering better results in both content preservation and the accuracy of applied edits, particularly in more complex tasks such as object substitution, background editing, and style transfer.}
    \label{fig:exp:edit:comparision}
\end{figure*}


\textbf{Quantitative Results} To evaluate image editing capabilities, we utilize the PIE-Bench dataset \cite{ju2024pnp}. This dataset includes 700 images and encompasses 10 distinct editing tasks.
As shown in Table \ref{tab:editing}, ABM-Solver consistently outperforms other methods, demonstrating its ability to maintain high preservation quality while aligning with the intended modifications.
ABM-Solver achieves efficient, high-quality edits with fewer steps by combining the multi-step ABM predictor-corrector method, which balances accuracy and speed. The Mask Guided Feature Injection module further enhances editing by preserving non-edited regions while allowing effective semantic changes.
Experiments on high-resolution image datasets confirm that ABM-Solver improves inversion precision and editing quality, outperforming existing solvers without additional training. Overall, it offers an optimal balance between accuracy and computational efficiency, making it highly effective for both precise and efficient editing tasks.

\textbf{Quantitative Results}
The visual results of image editing using the ABM-Solver are presented in Figure \ref{fig:exp:edit:comparision}.  Our approach effectively balances content preservation with the accuracy of applied edits, excelling in appearance editing, style editing, and geometry editing.
In appearance editing tasks, our ABM-Solver outperforms other methods. It can precisely modify specific elements while maintaining the integrity of surrounding areas. In contrast, methods like P2P, Pix2Pix-Zero, MasaCtrl, and PnP often introduce visual inconsistencies in unchanged regions.
Regarding style editing, ABM-Solver smoothly applies the desired style changes, such as converting to “Origami-Style” or “Oil Painting Style”, while preserving the original image’s overall structure. Others may disrupt the structure during the process.
For geometry editing that requires keeping the object’s identity, ABM-Solver strikes a better balance. Although methods like RF-Solver and FireFlow still have minor issues in unmodified regions. Meanwhile, Diffusion-based methods struggle to fully implement intended geometric changes.
Overall, ABM-Solver consistently delivers more accurate results across diverse editing prompts, making it a more reliable option for image editing, excelling in both content preservation and high-quality edits.

\subsection{Inversion and Reconstruction}
We conduct experiments on inversion and reconstruction tasks, comparing our ABM-Solver with the vanilla RF sampler, RF Solver, FireFlow  and DDIM Inversion.


\begin{table}[t]
    \centering
    \footnotesize 
    \caption{
    Quantitative comparison of inversion and reconstruction performance on DCI dataset \cite{dci_10658378}. ReFlow-Inv., RF-Solver, FireFlow, and ABM-Solver are based on the FLUX-dev model. The NFE includes both inversion and reconstruction function evaluations, and computational costs (steps) are kept comparable across methods.
    }
    \label{tab:inv_rec}
    \scriptsize
    \begin{tabular}{l c c c c c}
        \toprule
         & \textbf{Steps} & \textbf{NFE}$\downarrow$ & \textbf{LPIPS}$\downarrow$ & \textbf{SSIM}$\uparrow$ & \textbf{PSNR}$\uparrow$ \\
        \midrule
        DDIM-Inv. & 50 & 100 & 0.2342 & 0.5872 & 19.72 \\
        \midrule
        RF-Inversion \cite{rout2024semantic} & 30 & 60 & 0.5044 & 0.5632 & 16.57 \\
        RF-Solver \cite{wang2024taming}& 20 & 80 & 0.1653 & 0.5903 & 20.16 \\
        FireFlow \cite{deng2024fireflow} & 20 & 42 & 0.1359 & \textbf{0.6172} & {20.37} \\ 
        \textit{Ours (ABM-Solver)} & 15 & 40-60 & \textbf{0.1309} & \underline{0.6004} & \textbf{20.66} \\ 
        \bottomrule
    \end{tabular}
    \vspace{-3pt}

\end{table}

\begin{figure}[t]
    \centering
    \includegraphics[width=\linewidth]{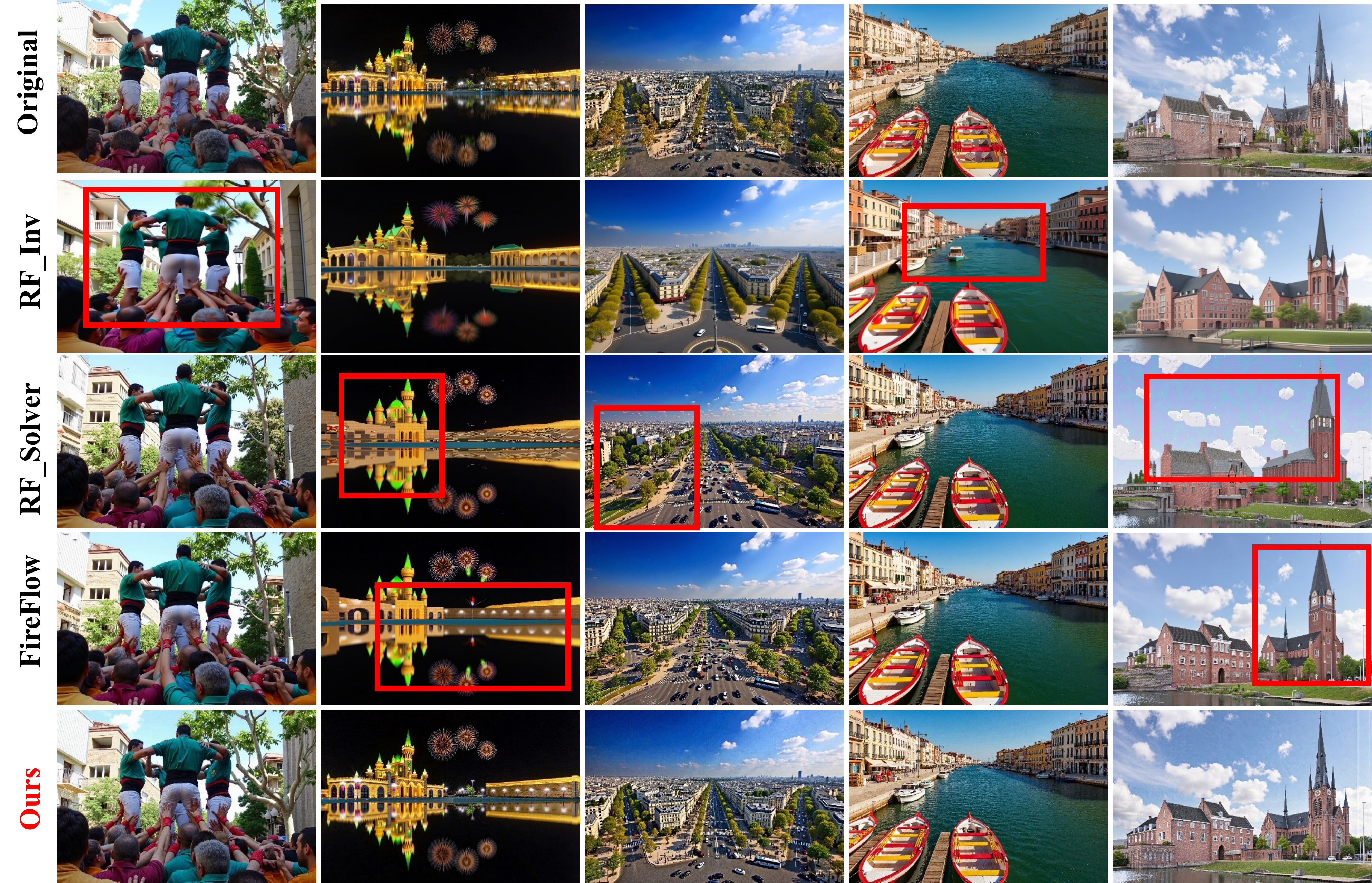}
    \caption{Qualitative results of image reconstruction with our ABM-Solver, FireFlow, RF-Solver, and ReFlow-Inv. The results demonstrate that ABM-Solver provides superior structural consistency and detail preservation compared to other methods. }
    \label{fig:exp:inversion:comparision}
    \vspace{-3pt}
\end{figure}

\textbf{Quantitative Results}
We presented the quantitative results of the Reconstruction Experiment in Table~\ref{tab:inv_rec}, using the DCI dataset \cite{dci_10658378}. It demonstrates that ABM-Solver outperforms the other methods across all four metrics. Despite slightly lower performance in one metric, our approach requires fewer steps to achieve comparable or superior image reconstruction quality, illustrating its efficiency and effectiveness. While FireFlow achieves slightly better SSIM scores, ABM-Solver requires only 15 steps compared to FireFlow’s 20, demonstrating its superior computational efficiency without sacrificing reconstruction quality.

\textbf{Performance Analysis}
ABM-Solver significantly reduces errors in solving the RF ODE, enhancing the accuracy of image reconstruction. As illustrated in Figure~\ref{fig:exp:inversion:comparision}, when the standard rectified flow approach is used, the reconstructed images exhibit notable deviations from the original, causing undesirable alterations in subject appearance. In contrast, ABM-Solver minimizes these deviations, providing higher fidelity reconstructions with fewer steps.

\begin{table*}[t]
\centering
    \caption{
    Ablation Study of Mask Guided Feature Injection in ABM-Solver
    }
    \begin{tabular}{l|ccccc}
    \toprule
    Methods & Distance$\downarrow$ & PSNR$\uparrow$ & SSIM$\uparrow$ & CLIP (Whole)$\uparrow$ & CLIP (Edit)$\uparrow$ \\
    \midrule
    w/o MGFI  & 0.0274 & 23.08 & 0.8142 & 24.98 & 22.01  \\
    \multicolumn{1}{l|}{\hspace{1em}– w/o Mask}   & 0.0271 & 23.65 & 0.8273 & 25.39 & 22.42 \\
    \midrule
    \textit{Ours} & \textbf{0.0207} & \textbf{24.60} & \textbf{0.8305} & \textbf{27.31} & \textbf{22.97}   \\
    \bottomrule
    \end{tabular}

\label{tab:ablation:mgfi}

\end{table*}

\begin{figure}[t]
    \vspace{-3pt}

    \centering

\includegraphics[width=\linewidth]{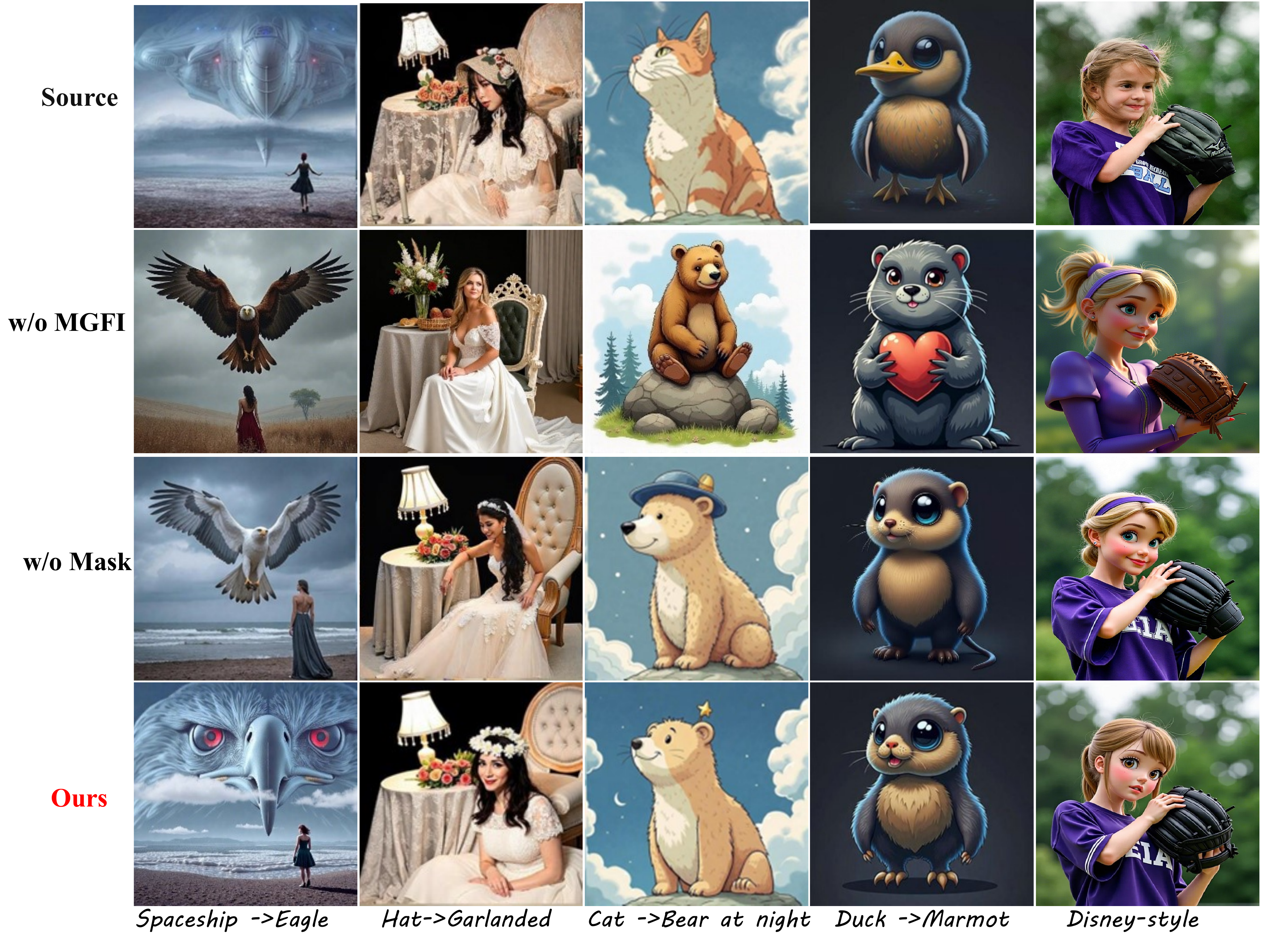}
    \caption{Ablation Results of Mask Guided Feature Injection module in our ABM solver.}
    \vspace{-12pt}
    \label{fig:exp:edit:ab}
\end{figure}

\subsection{Ablation Study}
\vspace{-3pt}
In this section, we conduct ablation studies to evaluate the effectiveness of the Mask Guided Feature Injection (MGFI) module in our proposed ABM-Solver. We investigate the impact of  different components of our method, including the inclusion of the mask and the feature Injection strategy. This analysis helps quantify the contribution of each component to the overall performance editing quality. As shown in Table~\ref{tab:ablation:mgfi}, the quantitative results clearly demonstrate the significant impact of the MGFI module. MGFI effectively improves editing quality and maintains background consistency. The qualitative results, presented in Figure~\ref{fig:exp:edit:ab}, further highlight that without the MGFI module, the edited images fail to maintain consistency between the original and edited content. Additionally, removing the spatial mask leads to noticeable deviations in certain fine details.

\begin{figure*}[t]
    \centering
\includegraphics[width=\linewidth]{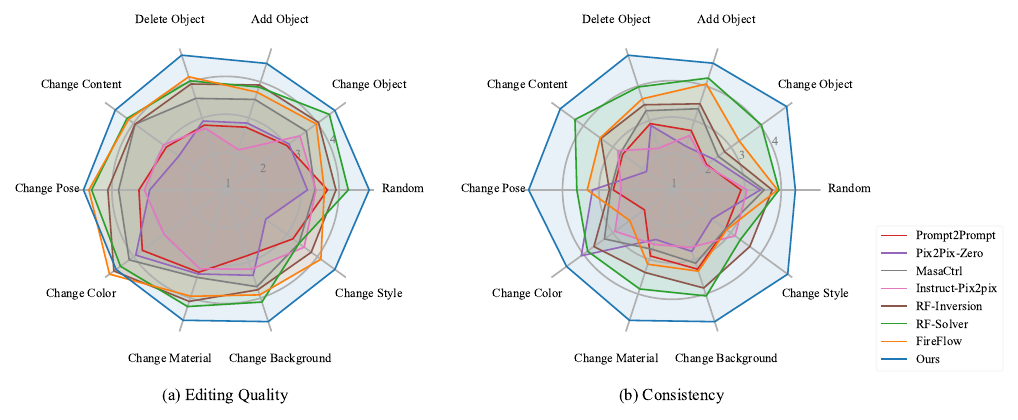}
    \caption{User study results comparing our method with several baseline approaches across various editing modalities. The study evaluates both Editing Quality (assessing visual fidelity, sharpness of the edited regions, and alignment with user prompts) and Consistency (measuring the coherence of the edited image with respect to the original, including aspects like color harmony, object placement, and semantic context). Paricipants (15 in total) rated the results of our method and baseline methods on a scale from 1 to 5 for each dimension. }
    \label{fig:userstudy}
        \vspace{-6pt}

\end{figure*}

\subsubsection{User Study}
To further evaluate the effectiveness of our method, we conducted a comprehensive user study. We considered several editing modalities, namely Change Material, Change Background, Change Style, Change Color, Change Pose, Change Content, Delete Object, Add Object, Change Object, and Random. For each modality, we presented two sub-figures: (1) focusing on Editing Quality, which was evaluated based on factors such as visual fidelity, sharpness of the edited regions, and how well the changes adhered to the user's intended prompt; and (2) addressing Consistency, which measured the overall coherence of the edited image with respect to the original, including aspects like color harmony, object placement, and semantic context.
Participants were asked to rate the results of our method and several baseline methods on a scale from 1 to 5 for each of these dimensions. The aggregated results, as shown in Figure~\ref{fig:userstudy}, clearly indicate that our approach outperforms the competitors in most editing modalities. In particular, for complex edits like Change Content and Add Object, our method received significantly higher ratings in both Editing Quality and Consistency, demonstrating its superiority in handling challenging editing tasks while maintaining a high level of user satisfaction.

\begin{figure*}[t]
    \centering
\includegraphics[width=\linewidth]{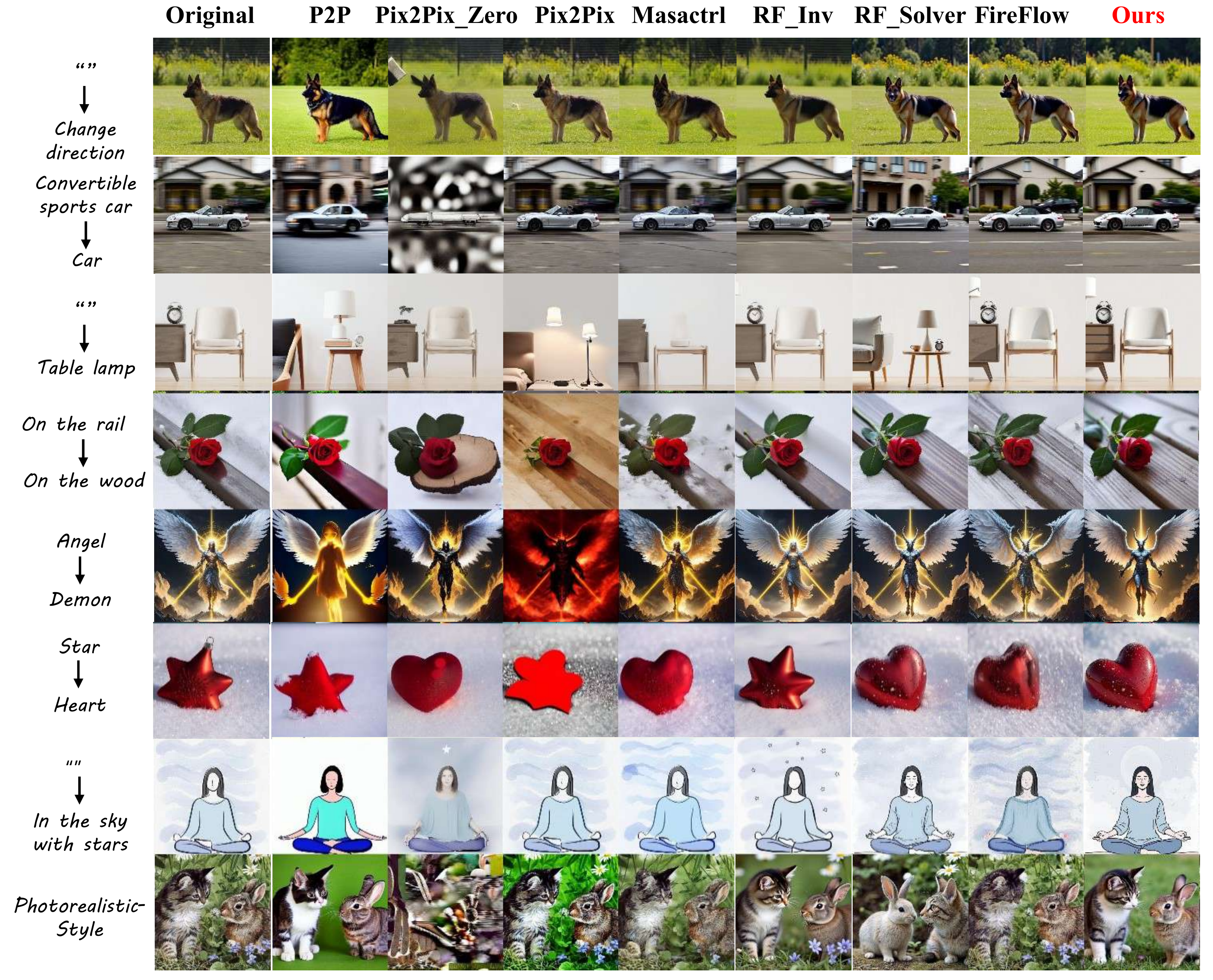}
    \caption{More qualitative comparison of image editing. }
    \label{fig:ablation:case:comp}
        \vspace{-6pt}

\end{figure*}

\begin{figure*}[t]
    \centering
\includegraphics[width=1\linewidth]{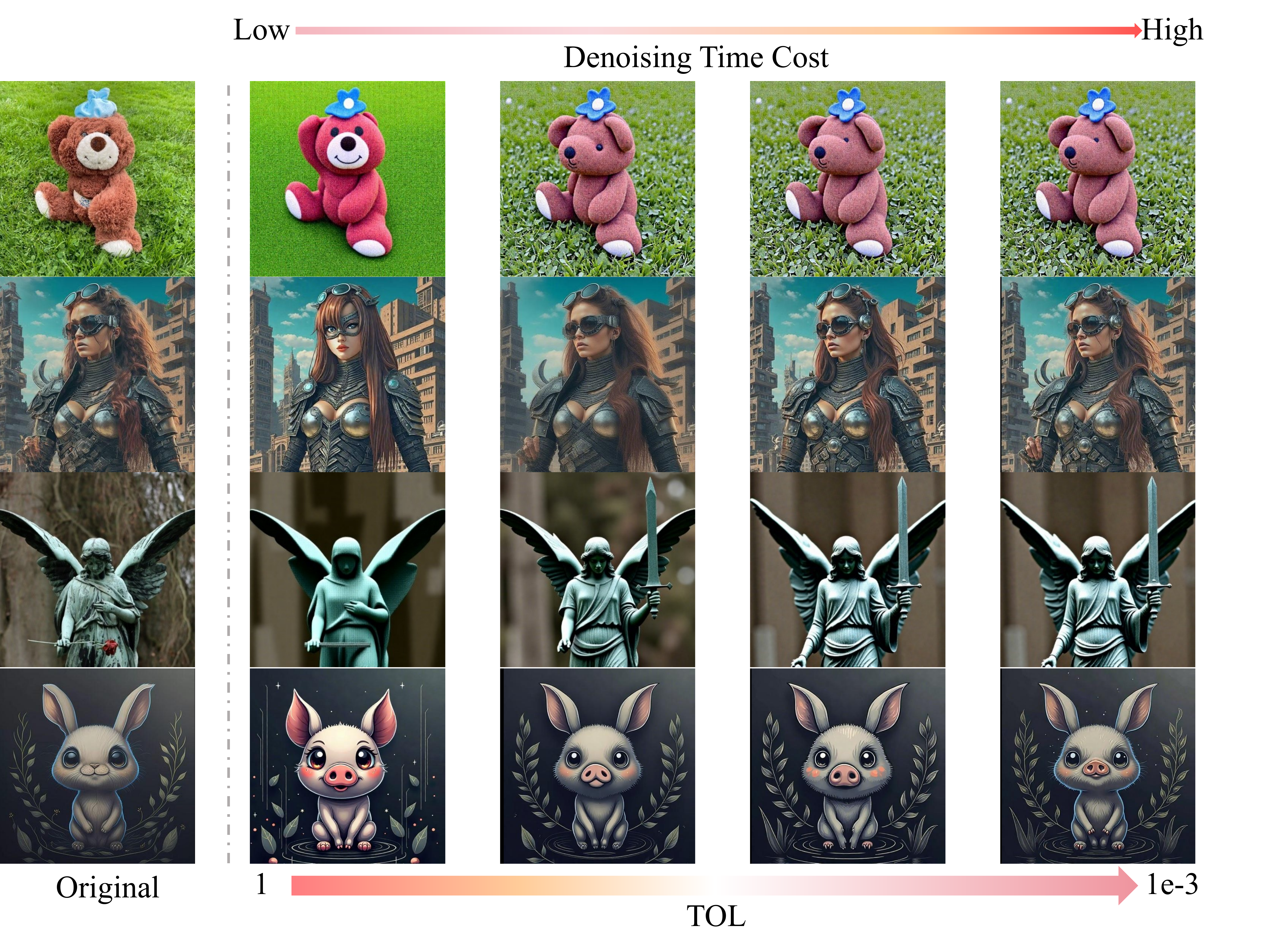}
    \caption{Qualitative results of the error tolerance $\epsilon$ experiment for the ASSA module. }
    \label{fig:ablation:TOL}
\end{figure*}

\section{Conclusion and Future Work}

In this work, we present ABM-Solver, a training-free, high-order numerical solver that leverages the Adams–Bashforth–Moulton predictor–corrector approach in conjunction with Adaptive Step Size Adjustment. By dynamically adjusting the integration steps, ABM-Solver effectively mitigates the accumulation of discretization errors while maintaining computational efficiency. Furthermore, by coupling it with our proposed Mask Guided Feature Injection module, ABM-Solver achieves precise inversion and high-fidelity, semantically guided editing. Extensive experiments on high-resolution datasets demonstrate that ABM-Solver not only significantly improves inversion accuracy but also surpasses existing solvers in both speed and fidelity. These advancements underscore the potential of rectified flow models for reliable, high-quality content generation and editing, paving the way for broader adoption across computational media and related fields.

\begin{figure}[t]
        \centering
    \includegraphics[width=\linewidth]{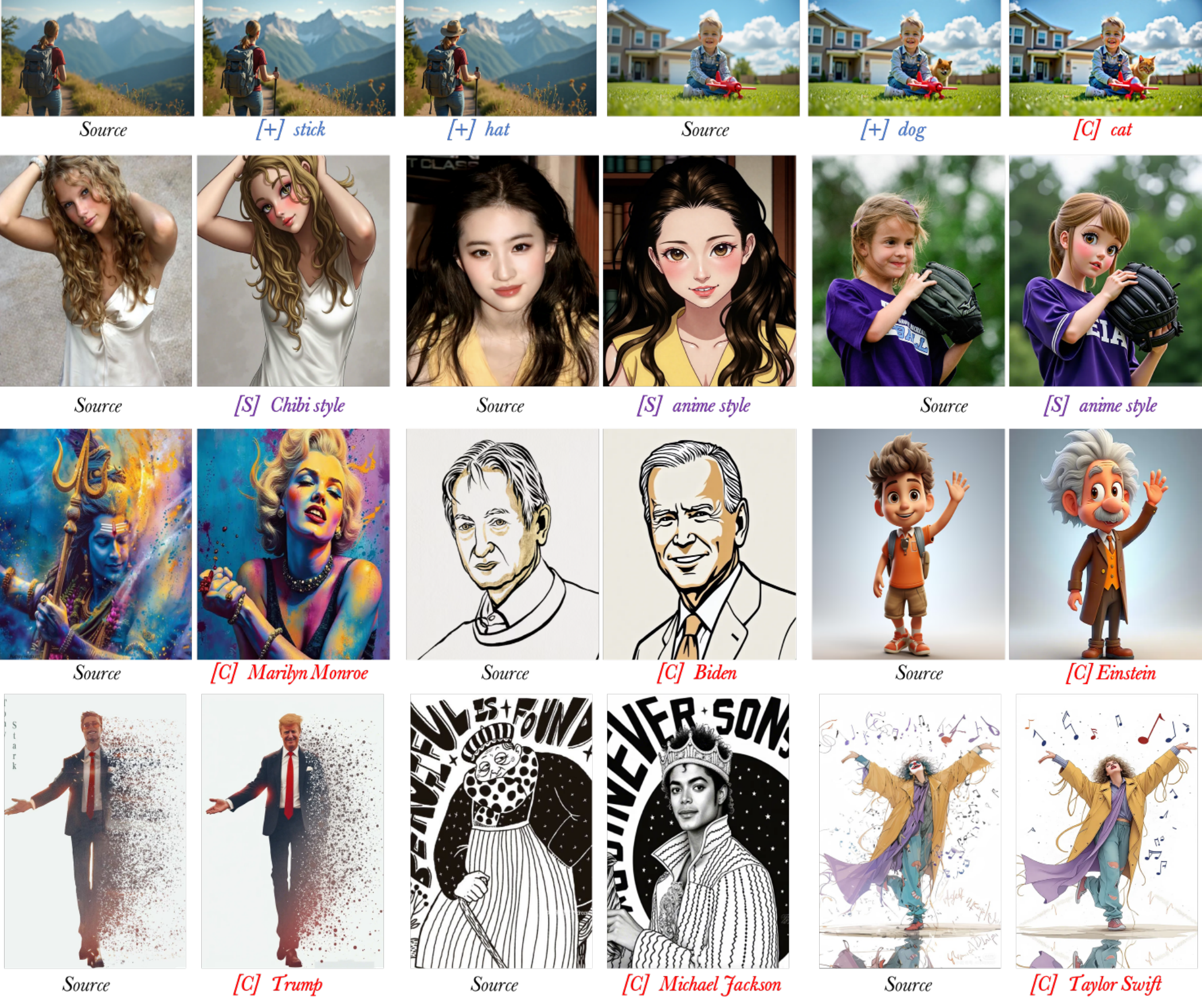}
                \vspace{-6pt}

       \caption{{Our approach achieves outstanding results in semantic image editing and stylization guided by prompts, while maintaining the integrity of the reference content image and avoiding undesired alterations. [+] means adding contents, [C] indicates changes in content and [S] denotes style editing.}
       }  
    \label{fig:teaser}

\end{figure}








\section*{Acknowledge}

    This work was supported by the National Key Research and Development Program of China (Grant No. GG-2024-01-02), National Natural Science Foundation of China (Grant No. 62277011) and Project of Chongqing MEITC(Grant No. YJX-2025001001009).

{
    \small
    \bibliographystyle{ieeenat_fullname}
    \bibliography{sn-bibliography}
}

\newpage

\section*{Appendix}

\label{sec:appendix}

To further validate the effectiveness of the ABM-Solver, we present additional experimental results in Figure~\ref{fig:ablation:case:comp} , comparing it to other image and video editing methods. These results reinforce our earlier findings, demonstrating that ABM-Solver provides superior performance in preserving non-edited regions while achieving high-quality, semantically consistent edits. In particular, the method excels in handling challenging edits, such as changing the main subject of an image or adding new objects, without introducing artifacts or inconsistencies. Moreover, global changes like style transfer are successfully applied with minimal disruption to the original content.

In contrast, baseline methods, such as P2P, Pix2Pix-Zero, MasaCtrl, and PnP, often produce unsatisfactory edits in certain areas, as seen in the additional visual comparisons. These methods struggle with maintaining the integrity of the original image in regions that should remain unchanged. While rectified flow-based models like RF-Inversion and RF-Solver perform better than the aforementioned methods, they still exhibit minor inconsistencies, particularly in preserving the original content and in areas requiring precise adjustments.

The ABM-Solver, however, consistently provides a more reliable and robust solution, offering better control over both the preservation of original content and the fidelity of the applied edits.

\subsection{Adaptive Step Size Adjustment}
In this experiment, we evaluate the effect of varying error tolerance $\epsilon$ values in the ASSA module, which controls the step size and, consequently, the trade-off between computational efficiency and precision. We conduct qualitative assessments for $\epsilon$ values ranging from \(1 \times 10^{-1}\) to \(1 \times 10^{-4}\), with the default value being \(1 \times 10^{-1}\).

The results demonstrate a clear trend: as the $\epsilon$ value increases (\ie, allowing for larger error tolerance) in Figure~\ref{fig:ablation:TOL}, the step size increases, leading to faster progress through the process. However, this comes at the cost of greater approximation errors, which are visible in the reconstructed images and edits. Specifically, higher $\epsilon$ values result in noticeable artifacts, such as blurriness in the edited regions, misalignment of objects, and inconsistencies in color harmony. This degradation in performance is most apparent in the complex editing tasks, where precision is crucial for maintaining the fidelity of both the original content and the desired modifications.

On the other hand, as the $\epsilon$ value decreases (\ie, tighter error tolerance), the step size becomes smaller, leading to slower but more precise processing. While this results in more accurate reconstructions and edits, it requires more computational resources, making the process slower overall.

In terms of the ASSA mechanism, we introduce an important operational detail: the error tolerance adjustment does not affect the step size during the first 5 steps and the last 5 steps of the process. This is done to maintain stable initial and final stages of the computation. During these steps, the method ensures that the step size aligns with the original second-order ABM solver, preserving consistency with the initial numerical solver.

Additionally, the minimum step size is set to be consistent with the second-order ABM sampling, ensuring that no steps are too small to cause unnecessary computational overhead. The maximum allowable step size is constrained to 4 times the initial step size, ensuring that the adjustment mechanism does not excessively accelerate the process and result in larger errors. This means that, for a total of 15 steps, the theoretical number of function evaluations (NFEs) falls within the range of 40 to 60, depending on the chosen $\epsilon$ value.

Our experiment results reveal a trade-off between speed and accuracy. Larger $\epsilon$ values may speed up the process but at the cost of increased errors and lower editing quality. Conversely, smaller $\epsilon$ values equate to a more precise method but at the cost of computational efficiency, often resulting in a process that is closer to the original second-order ABM method without utilizing the ASSA step adjustment


\subsection{Image Editing}
It demonstrates the effectiveness of our method in semantic image editing and stylization tasks in 
Figure~\ref{fig:teaser}. The images show how our approach preserves the core structure and content of the reference image while enabling modifications as specified by the input prompts. Notably, when content is added [+], the existing structure remains intact without undesired alterations. When content is changed [C], our model carefully adjusts the relevant regions, ensuring that the edits are both natural and semantically consistent with the original image. Style editing [S] also reveals the model's ability to modify the overall style of the image while preserving the underlying content. This illustrates the high fidelity and flexibility of our approach in various semantic editing scenarios.

\section{Theoretical Analysis}

\subsection{Error Analysis of the Second-Order ABM Solver}
\label{proof:abm_error}

\begin{proof}[Sketch of the Local and Global Error Bounds]
Again consider the ODE
\begin{equation}
   \frac{dZ}{dt} \;=\; f(Z, t),
   \quad Z(0)=Z_0,    
\end{equation}
on a time interval $t \in [0,T]$.  Let $h = \max_i (t_{i+1}-t_i)$ denote the largest step size. We show that the local truncation error is $\mathcal{O}(h^3)$, which in turn implies a global error $\mathcal{O}(h^2)$.

\paragraph{(a) Local truncation error}
The local truncation error (LTE) measures the discrepancy if one step of the ABM update is applied using exact solution values at the previous steps.  For the two-step Adams-Bashforth predictor, one can expand $f(Z(t), t)$ in Taylor series about $t_i$, while for the Adams-Moulton corrector, a similar expansion centers on the interval $[t_i, t_{i+1}]$.  Standard polynomial interpolation arguments (or standard texts in numerical analysis) show that for sufficiently smooth $f$:
\begin{equation}
   \text{LTE}_{\text{ABM}} \;=\; \mathcal{O}(h^3).
\end{equation}
\vspace{-6pt}
\paragraph{(b) Global error}
By consistency and stability arguments (\eg, a discrete Grönwall lemma), the $\mathcal{O}(h^3)$ local error guarantees that the global accumulated error remains $\mathcal{O}(h^2)$ over the full interval $[0,T]$.  Formally, let $Z(t)$ be the exact solution and $Z_{t_i}^{(\mathrm{ABM})}$ the numerical solution at discrete steps.  Then there exists a constant $C'$ (independent of $h$) such that 
\begin{equation}
   \max_{i}\,\bigl\|\,Z(t_i) \;-\; Z_{t_i}^{(\mathrm{ABM})}\bigr\| 
   \;\le\; C'\,h^2,
\end{equation}
provided $f$ is sufficiently smooth and Lipschitz continuous in $Z$.

Thus, the second-order ABM solver enjoys $\mathcal{O}(h^3)$ local truncation error and $\mathcal{O}(h^2)$ global convergence, as stated in the main text.
\end{proof}

\subsection{Adaptive Step Size Preserves the Convergence Order}
\label{proof:adapt_step_size}

\begin{proof}
Let $\{t_0, t_1, \ldots, t_N\}$ be nonuniform points in $[0,T]$ determined by an adaptive step size strategy.  Denote $h_i = t_{i+1}-t_i$ and $h_\mathrm{max} = \max_i h_i$.  The method’s local error control (involving a tolerance $\varepsilon$) ensures that each individual step’s local truncation error satisfies
\begin{equation}
   \|\text{LTE}_i\| \;\le\; K\,h_i^{p+1},
\end{equation}
for some constant $K$ and method order $p=2$.  Whenever the error estimate $E_{i+1}$ in Equation~\eqref{eq:error_estimate} exceeds the threshold $\varepsilon$, the step is rejected and $h_i$ is decreased (see Equation~\eqref{eq:step_adapt}).  Conversely, if $E_{i+1}$ is well below $\varepsilon$, one may increase $h_i$.

Because each successful step $i$ still adheres to 
\begin{equation}
   \|\text{LTE}_i\| \;\le\; C\,h_i^{3}, \quad (p=2),
\end{equation}
the standard global convergence argument remains valid on each subinterval $[t_i, t_{i+1}]$.  As long as $h_\mathrm{max} \to 0$ with $N \to \infty$, the global error bound on $[0,T]$ follows the same $\mathcal{O}(h_\mathrm{max}^2)$ rate as in the uniform-step case.  Informally:
\begin{equation}
   \max_{0 \le i \le N} \|\,Z(t_i) \;-\; Z_{t_i}^{(\mathrm{ABM})}\|
   \;\le\; C'\,\bigl(h_\mathrm{max}\bigr)^2,
\end{equation}
where $C'$ depends on the Lipschitz constant of $f$ and the chosen error tolerance $\varepsilon$, but not on $h_\mathrm{max}$ itself.  This shows that adaptive step size control, when combined with an error-estimation step consistent with the order of our ABM solver, preserves the second-order global convergence.
\end{proof}

\end{document}